\documentclass[runningheads]{llncs}
\usepackage{graphicx}
\usepackage{mathrsfs}
\usepackage{algorithm}
\usepackage{algorithmic}
\begin{document}
\title{Using Sentiment Representation Learning to Enhance Gender Classification for User Profiling \\
%\thanks{Supported by organization x.}
}

\author{Yunpei Zheng\inst{1} \and
Lin Li\inst{1} \and
Luo Zhong\inst{1} \and
Jianwei Zhang\inst{2} \and
Jinhang Liu\inst{1}}
\authorrunning{Y. Zheng et al.} % First names are abbreviated in the running head. If there are more than two authors, 'et al.' is used.
\institute{Wuhan University of Technology, Wuhan, China \and
Iwate University, Japan\\
\email{PPgirl87@foxmail.com,}\\
\email{cathylilin@whut.edu.cn,}\\
\email{zhongluo@whut.edu.cn,}\\
\email{zhang@iwate-u.ac.jp,}\\
\email{ryukinkou@whut.edu.cn.}}

\maketitle              % typeset the header of the contribution
\begin{abstract}
User profiling means exploiting the technology of machine learning to predict attributes of users, such as demographic attributes, hobby attributes, preference attributes, etc. It's a powerful data support of precision marketing. Existing methods mainly study network behavior, personal preferences, post texts to build user profile. Through our data analysis of micro-blog, we find that females show more positive and have richer emotions than males in online social platform. This difference is very conducive to the distinction between genders.  Therefore, we argue that sentiment context is important as well for user profiling.This paper focuses on exploiting microblog user posts to predict one of the demographic labels: gender. We propose a Sentiment Representation Learning based Multi-Layer Perceptron(SRL-MLP) model to classify gender. First we build a sentiment polarity classifier in advance by training Long Short-Term Memory(LSTM) model on e-commerce review corpus. Next we transfer sentiment representation to a basic MLP network. Last we conduct experiments on gender classification by sentiment representation. Experimental results show that our approach can improve gender classification accuracy by 5.53\%, from 84.20\% to 89.73\%.

\end{abstract}
\section{Introduction}
User profiling is a labeled user model abstracted from information like user social attributes, lifestyle and consumer behavior. The key work of building a user profile is to label users with some highly refined features that can summarize user characteristics through analysis of various user information, in a word, digitizing users. User profiling has multiple applications, such as precision marketing. User profiling can help analyze potential users of products and market them by sending messages or emails. It can also be applied to data statistics and decision support, helping have more detailed understanding of users to develop personalized service or offer higher level of service for retaining users or supporting product transformation. Gender label helps to design a personalized product, filter out products that don't match gender, recommend contents for a certain gender, etc. But in social media, filling the basic information is not compulsive. As a result, user information can be missing or untrue, and situations also exist that user preference does not match the actual gender. So it is necessary to predict the gender of the user.

Centering on user profiling, researchers utilize various information to predict target tasks. From posts or self-descriptions, many use lexical features\cite{aaai_short}, textual content features\cite{classify_occup} or else. From other information, features will be extracted from images, following relations, hyperlinks, consumptions\cite{WWW_topical_interests} and other behaviors.  Some experiments have been conducted to predict gender\cite{discriminate_gender_480}, age, occupation, education, interest or other dynamic attributes. As for sentiment analysis, sentiment analysis can be applied to public opinion analysis and political tendency analysis. It can even help predict stock price movements\cite{EMNLP_stock}\cite{ACL_stock} and monitor bank risks\cite{EMNLP_bank}. In this paper, we study gender classification for micro-blog users. We consider introducing sentiment representation to enhance gender prediction.

\begin{figure}[t]
\centerline{\includegraphics[width=7.8cm,height=5.3cm]{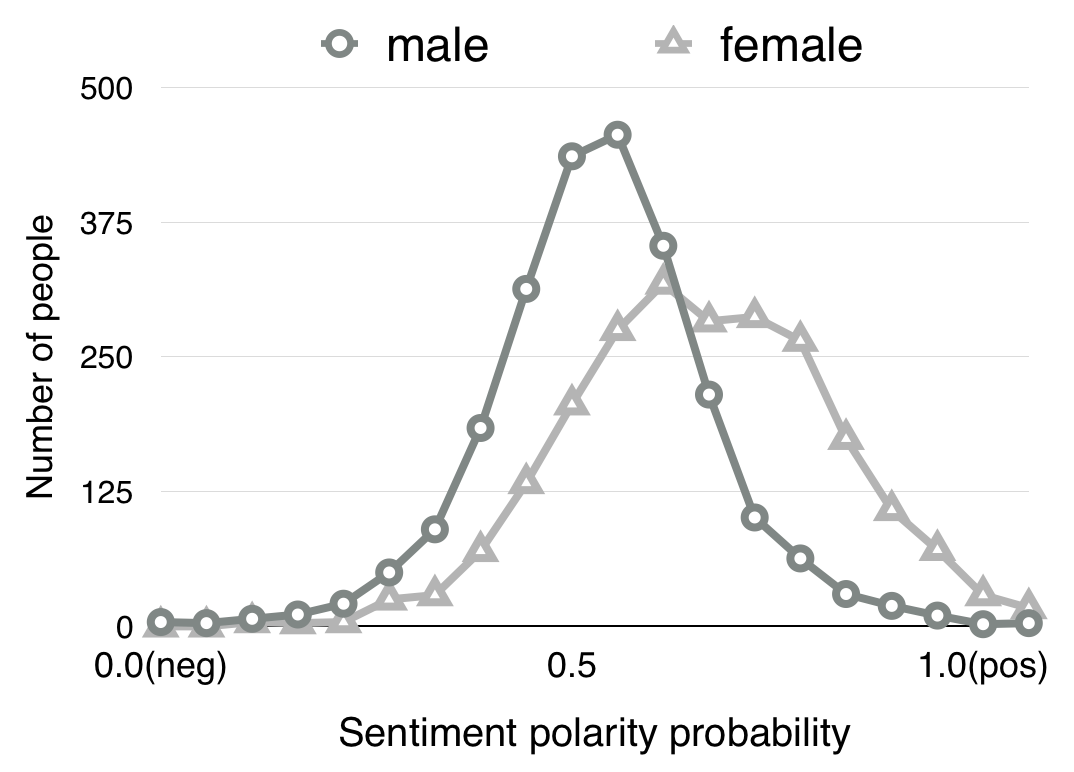}}
\caption{Sentiment polarity probability difference between male and female.}
\label{sentiment_difference}
\end{figure}

In order to improve the accuracy of gender classification, it is very important to select effective features. Features determine the upper bound, and models are to keep approaching the upper bound. After investigating the data, we find there exits sentiment difference between male and female. For example, after investigating 3138 users' posts, for the same topic football, word \emph{football} in Chinese occurs 646 times in male, while 39 times in female, and their feelings about football are quite different. Most attitudes of male to Chinese football are very negative, while female are indifferent and even a little positive. In order to make a intuitive representation of the sentiment difference between male and female, we make a sentiment polarity probability distribution of them, we can see from Fig.~\ref{sentiment_difference} that, most of the micro-blog posts sent by male are very neutral, while on the whole, female are more positive than male. Maybe female always like to show good mood, but male only send a plain post when they encounter a big events. And we can see from the figure that female have larger sentiment span than male.

As a consequence, we decide to add sentiment features to improve model's gender classification performance. However, there exist two questions:
\begin{itemize}
\item How to get sentiment labels in micro-blog training sets, since there is no sentiment information in micro-blog.
\item How to effectively represent sentiment.
\end{itemize}
Our main contribution is listed as follows to solve the above questions:
\begin{itemize}
\item In order to get sentiment representation of micro-blog, we train a LSTM sentiment classifier from source domain of e-commerce reviews whose data are selected by calculating the similarity with micro-blog target domain.
\item We input micro-blog posts to LSTM to get target domain sentiment representation. In this paper we get frozen lstm layer's output to be our sentiment representation. Then we combine micro-blog post vectors and sentiment representation to form concatenated features.
\end{itemize}

In addition, since there is imbalanced data, we use smote oversampling\cite{smote} to adjust the train set. We conduct experiments on micro-blog, and experimental results show that our method can improve gender classification by 5.53\%. The rest of this paper is organized as follows: Section 2 reviews the related work of user profiling and sentiment analysis, section 3 describes our approach in detail, section 4 validates our proposal using experiment results, and section 5 makes a conclusion of our work and discusses our future work.

\section{Related Works}
\subsection{User Profiling}
User profiling has attracted much research efforts. Volkova et al.\cite{aaai_short} learn models to infer various traits from user communications in social media. They use crowdsourcing to annotate user profiles and train log-linear models using lexical features. Farnadi et al.\cite{WSDM} merge multiple modalities of user data, such as text, images and relations, to predict age, gender, and personality. They build a hybrid user profiling framework which utilizes a shared representation between modalities to integrate multiple sources of data at feature level and decision level. Mutliview based analysis is another effective way to deal with multiple modal data.\cite{YangW1,YangW2,YangW3}. Preotiuc-Pietro et al.\cite{classify_occup} conduct an analysis on a new annotated corpus, their posted textual content or else, to predict the occupational class. They employ non-linear methods using latent feature representations, and get strong accuracy. Burger et al.\cite{discriminate_gender_480} construct a large, multilingual dataset labeled with gender, and they mainly use screen name, full name, description, tweets to predict gender. Zhao et al.\cite{WWW_topical_interests} build personal topic interest profiles by analysing consumption and publishing behaviors, and they propose to separately model users' topical interests that come from various behavioral signals to improve. Chao et al.\cite{SIGIR_footprint} propose a model that simultaneously considers multiple footprints to build user topical profiles. Researchers have exploited wide various forms of data, and natural language processing researchers also have a deep research on lexical and syntactic features. However, to the best of our knowledge, few researchers take sentiment into account when building user profiles.

\subsection{Sentiment Analysis}
For sentiment analysis, Li et al.\cite{cross_domain_transfer_sentiment} proposes a Hierarchical Attention Transfer Network(HATN) for cross-domain sentiment classification. Wang et al. \cite{capsule_sentiment_analysis} build a RNN-Capsule for sentiment analysis without using any linguistic knowledge, and they get very good results on movie reviews and other datasets. Si et al.\cite{EMNLP_stock} explore stock prediction, and they utilize close neighbors' topic-sentiment time-series to help. Nguyen et al.\cite{ACL_stock} find a new topic-sentiment feature and build a TSLDA model which can capture topics and sentiment on social media simultaneously to improve predicting stock price movement. Wang et al.\cite{ACL_dispute} investigate a novel task of online dispute detection, and propose to identify the sequence of sentence-level sentiment expression to predict dispute/non-dispute labels. Nopp et al.\cite{EMNLP_bank} reveal significant correlations between risk sentiment analysis and macroprudential analysis in the banking system. Pla et al.\cite{COLING_political} develop a sentiment analysis system to detect user political tendency. Various researchers have developed excellent attention models for sentiment analysis. Sentiment analysis has so many applications such as public opinion analysis, political tendency analyais, information prediction, however, to the best of our knowledge, there are few researchers who apply it to user profiling.

\section{Our SRL-MLP Approach}

\subsection{Framework}

\begin{figure}[h]
\centerline{\includegraphics[width=12cm,height=5.9cm]{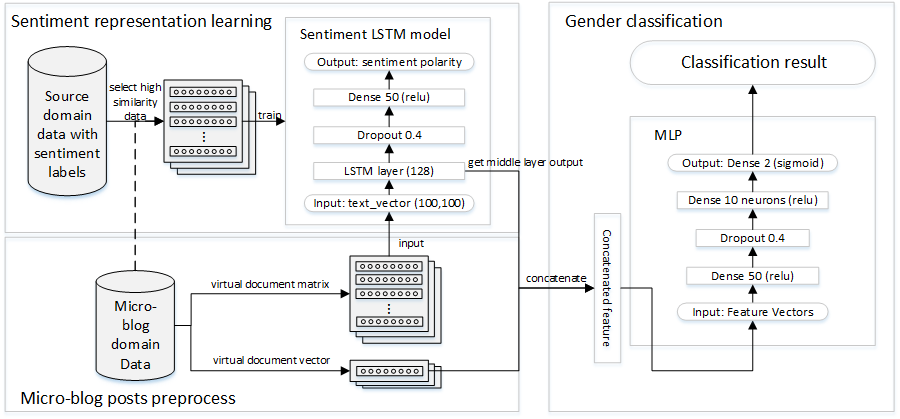}}
\caption{The framework of our SRL-MLP approach.}
\label{framework}
\end{figure}

This paper utilizes supervised approach to classify gender of micro-blog users. The overview of our proposed approach is shown in Fig.~\ref{framework}. This framework include three parts. The first part is micro-blog data preprocess. In this part we input segmented micro-blog posts, then we get the virtual document matrix and virtual document vector. What is virtual document will be illustrated in the following subsection of micro-blog posts preprocess. The second part is sentiment representation learning. In this part we train a sentiment analyzer on source domain and get sentiment representation of target domain data. The third part is gender classification. In this part we concatenate the virtual document vector in micro-blog posts preprocess part with the sentiment representation in sentiment representation learning, and train a gender classifier.

\noindent Since MLP can not effectively analyze the order of words, we consider some deep learning models, like Long Short-Term Memory(LSTM) and with Convolutional Neural Network(CNN), to get sentiment representation learning model. After comparing them in experiment section, we choose LSTM. The sentiment representation learning LSTM model we exploit is shown in the sentiment representation learning part of Fig.~\ref{framework}. Our LSTM model includes one lstm layer with dropout and a dense layer. We use Multi-Layer Perceptron(MLP) shown in the gender classification part of Fig.~\ref{framework} to classify gender. Our MLP architecture consists of two hidden layers of different neurons and one layer dropout, for too much parameters make training longer, and we don't have too much data. We have tried Random Forest and Support Vector Machine and MLP, but we find MLP with back propagation performs the best.

\subsection{Micro-blog Posts Preprocess}

\begin{figure}[h]
\centerline{\includegraphics[width=11cm,height=7cm]{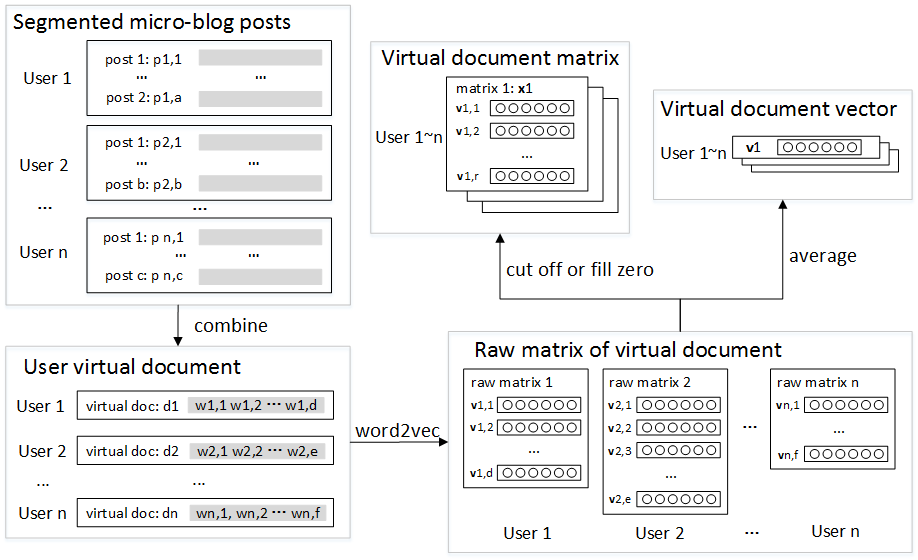}}
\caption{Micro-blog posts preprocess.}
\label{preprocess}
\end{figure}

In the previous section we mention the concept of virtual documents, in this part we illustrate our preprocess in detail. The flowchart of micro-blog posts preprocess is shown in Fig.~\ref{preprocess}. Users of micro-blog may post different numbers of posts, and the numbers of words in each post are also different. And micro-blog limits each post 140 words, for gender classification it is too short. We combine all the posts posted by the same user to form a virtual document, so that each user corresponds to one virtual document.

\noindent For LSTM and MLP receive different data input formation, we need to prepare two different forms of data. One is virtual document matrices which will be input to sentiment representation learning model to get sentiment representations. The another one is virtual document vectors that are used to be concatenated with the sentiment representations. After getting virtual documents, documents are vectorized to get raw matrices of virtual document. The reason why the matrices are raw is that the shape of these matrices are different due to different number of words. For forming a virtual document vector, we only need to add each vector in a raw matrix up and get their average. However, for the formation of a virtual document matrix, we need to cut off many words or fill zero vector. As shown in Fig.~\ref{preprocess}, we only take the first r words and zero-pad the virtual documents with more or less than r words, so that the shape of all virtual document matrices are the same.

\subsection{Sentiment Representation Learning}
Sentiment analysis is a hot topic, but it is extremely hard to get accurate labeled data in social media data. Although we can manually label some posts, it is unrealistic to get an accurate model through limited manual labels. Inspired by the idea of transfer learning, we can utilize labeled data from source domain that are close to the target domain data to build a sentiment analysis model for micro-blog, as shown in sentiment representation learning part in Fig.~\ref{framework}.  Unlike micro-blog, we are always pushed to comment a commodity with a short text with a rate 1 to 5 stars in commodity reviews. We can clearly get sentiment polarities, positive or negative, from these rate, except from some spam comments. So from commodity reviews domain, we can easily get documents with sentiment polarity labels. As a consequence, after getting a source domain trained model, we can transfer it to target domain to get sentiment representation.

\noindent\textbf{Select high similarity data}. Our current task is to learn sentiment from a source domain then transfer it into social media domain. Transfer learning leverages the similarity between source domain and target domain. Only when we find this similarity and make full use of it can we accomplish transfer learning, otherwise negative transfer will occur. The main reasons for negative transfer are data problems and methodological problems. Although we have selected the labeled data that are as close as possible to target domain, we still have other techniques to make source domain data and target domain data closer.

\noindent Authoritative review paper\cite{transfer_learning_survey} summarizes that the basic methods of transfer learning are instance based transfer, parameter based transfer, feature based transfer, and relation based transfer. In general, source domain $\mathcal{D}_{s}=\{\textbf{x}_{k},y_{k}\}_{k=1}^{m}$ and target domain $\mathcal{D}_{t}=\{\textbf{x}_{i}\}_{i=m+1}^{m+n}$ probability distributions are usually different and unknown. Based on instances, we can make source domain and target domain probability distributions even closer.

\noindent One method is that we can select source domain data which have high similarity with the target domain to be the new source domain training data. This method need similarity calculation, so we get the average vector $\textbf{v}_k,k=1,2,...,m.$ of each source domain data, and the average vector $\textbf{v}_i,i=1,2,...,n.$ of each virtual document vector. Then we calculate the average similarity between each $\textbf{v}_k$ and all $\textbf{v}_k$. Finally we select the corresponding source document matrix $\textbf{x}_k$, according to subscript $k$, whose average similarity exceed $z$, to be source domain data. Parameter $z$ is a similarity threshold. Equation is shown in Equation \ref{select_high_sim}.

\begin{equation}
\mathcal{D}_{s}\leftarrow\{\textbf{x}_{k},y_{k}|\frac{1}{n}\sum_{i=1}^{n} SIM(\textbf{v}_{k},\textbf{v}_{i})>z, z\in(0,1)\}
\label{select_high_sim}
\end{equation}

\noindent Another method is adding some manually labeled target domain data to source domain, and making the expanded data set to be new source domain data. As Equation \ref{manual_label} shown, $\textbf{x}_{mlt}$ is manually labeled sample in target domain.

\begin{equation}
\mathcal{D}_{s}\leftarrow\mathcal{D}_{s}\cup\{\textbf{x}_{mlt},y_{mlt}|\textbf{x}_{mlt}\in \mathcal{D}_{t}\}
\label{manual_label}
\end{equation}

\noindent\textbf{Sentiment representation.} After have trained a LSTM model with new source domain data which have been selected or expanded, we get a model that can predict the sentiment polarity. Yosinski et al.\cite{deep_transform} have validated by experiments that features are transferable in deep neural networks. Parameter based transfer learning tells us that some model parameters in the source and target domains can be shared. Now most of parameter based transfer learning methods are combined with deep neural network, and it is effective on different learning tasks.

\noindent Finetune is the simplest method of deep network transfer, which uses already trained network to fine-tune on specific tasks. It provides a good reference or auxiliary method for the traditional artificially extracting features method which may cost a lot. We can use deep neural networks to train and rely on the networks to extract richer and more expressive features. Extracted features are then used as input or part of input for traditional machine learning methods.

\noindent More specifically, after having trained a sentiment analysis model, for example we have got a LSTM model trained on e-commerce commodity reviews and we can input our micro-blog posts to this LSTM model to get micro-blog domain sentiment representation. Apart from the final result which tells sentiment polarity, we can get its middle layer outputs to be our sentiment representations.

\subsection{Gender Classification}

\begin{figure}
\begin{minipage}{0.55\linewidth}
\centerline{\includegraphics[width=7cm,height=5cm]{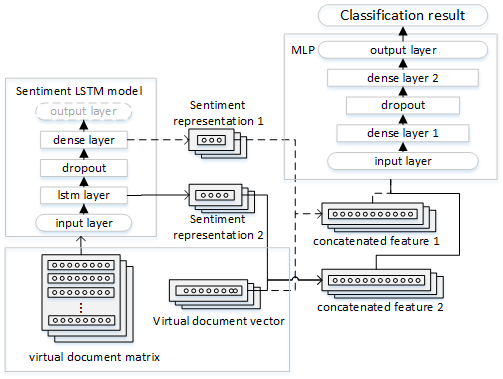}}
\centerline{(a) Add frozen layer.}
\end{minipage}
\hfill
\begin{minipage}{0.45\linewidth}
\centerline{\includegraphics[width=5cm,height=5cm]{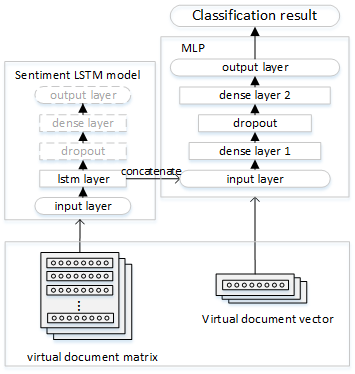}}
\centerline{(b) Add finetuned layer.}
\end{minipage}
\caption{Sentiment polarity probability difference between male and female.}
\label{mlp}
\end{figure}

In section 3.2 we get micro-blog post sentiment representations. There are many ways to concatenating sentiment representations and virtual document vectors, such as concatenating sentiment representation with virtual document vectors in the MLP input layer like sentiment representation 1 in Fig.~\ref{mlp}(a), or concatenating sentiment representation with MLP dense layer output or else. But through experiments, we find that directly concatenating in the input layer works the best. The preceding concatenating methods are frozen layer concatenation. For finetuning, we concatenate lstm layer of LSTM with MLP input layer to form a multi-input model, and finetune parts of origin LSTM model when training gender classifier like shown in Fig.~\ref{mlp}(b).

\subsection{Algorithm Description}
In this section, we give the algorithm of our approach. In target domain, micro-blog, let i represents user id and j represents user posts number. $p_{i,j}$ represents the post $j$ of user $i$, and $c_{i,j}$ is the count of words in $p_{i,j}$. In source domain, let $k$ represents commodity reviews number. $d_k$ represents document $k$ in source domain, and $c_k$ represents the word count of document $k$ in source domain. The final result is $G_{i}$ which represents the gender of user i. $d$ is the word vector dimension, and $r$ is the maximum count of words, so that each document matrix has the shape of $d\times r$, both micro-blog virtual document matrix and source domain document matrix. More details are shown in Algorithm \ref{algo}.

\renewcommand{\algorithmicrequire}{ \textbf{Input:}} %Use Input in the format of Algorithm
\renewcommand{\algorithmicensure}{ \textbf{Output:}} %Use Output in the format of Algorithm
\begin{algorithm}[h]
\caption{SRL-MLP}
\label{algo}
\begin{algorithmic}[1] %show procedure order number
\REQUIRE ~~\\ %Input
Segmented user micro-blog posts $p_{i,j}=\{w_{i,j}^{(1)},w_{i,j}^{(2)},...,w_{i,j}^{(c_{i,j})}\}$, i=1,2,...n;\\
Segmented source domain documents $d_{k}=\{w_{k}^{(1)},w_{k}^{(2)},...,w_{k}^{(c_{k})}\}$, k=1,2,...,m;
\ENSURE ~~\\ %Output
Output: Gender of user i, $G_{i}$.

\STATE Documents vector representation;\\
1.1 Assemble user posts to form user virtual document: \\
$d'_{i}=\{p_{i,1},p_{i,2},...,p_{i,j},...,p_{i,a}\}, i=1,2,...,n$;\\
Regardless of which post the word comes from, renumber the word: \\
$d_{i}=\{w_{i}^{(1)},w_{i}^{(2)},...,w_{i}^{(c_{i})}\}, i=1,2,...,n$;\\
1.2 Vectorize each word in virtual document to form raw matrix of document by word2vec\cite{w2v_1}\cite{w2v_2}, including micro-blog and source domain data. \\
$\textbf{x}'_{i}=\{\textbf{v}_{i}^{(1)},\textbf{v}_{i}^{(2)},...,\textbf{v}_{i}^{(c_{i})}\}, i=1,2,...,n$; $\textbf{x}'_{k}=\{\textbf{v}_{k}^{(1)},\textbf{v}_{k}^{(2)},...,\textbf{v}_{k}^{(c_{k})}\},k=1,2,...,m$;\\
1.3 Make average micro-blog word vectors:\\
$\textbf{v}_{i}=\frac{1}{c_{i}}(\textbf{v}_{i}^{(1)}+\textbf{v}_{i}^{(2)}+...+\textbf{v}_{i}^{c_i})$;
$\textbf{v}_{k}=\frac{1}{c_{k}}(\textbf{v}_{k}^{(1)}+\textbf{v}_{k}^{(2)}+...+\textbf{v}_{k}^{c_i})$;\\
1.4 Cut off or fill zero to make document matrices having same shape $d\times r$:\\
$\textbf{x}_{i}=\{\textbf{v}_{i}^{(1)},\textbf{v}_{i}^{(2)},...,\textbf{v}_{i}^{(r)}\}, i=1,2,...,n$; $\textbf{x}_{k}=\{\textbf{v}_{k}^{(1)},\textbf{v}_{k}^{(2)},...,\textbf{v}_{k}^{(r)}\},k=1,2,...,m$;\\
\label{algo:vectorize}

\STATE LSTM Sentiment representation learning;\\
2.1 Select high similarity source domain data to be new source domain data $\mathcal{D}_{s}$:\\
$\mathcal{D}_{s}\leftarrow\{\textbf{x}_{k},y_{k}|\frac{1}{n}\sum_{i=1}^{n} SIM(\textbf{x}_{i},\textbf{x}_{k})>a, a\in(0,1)\}$;\\
2.2 Using source domain data $\mathcal{D}_{s}$ to train a LSTM sentiment model.\\
2.3 Put target domain data $\mathcal{D}_{t}=\{x_{i}\},i=1,2,...,n$, into LSTM, for each $\textbf{x}_{i}$ get its lstm layer output $\textbf{h}_{i}$ to be our target domain sentiment representation:\\
\label{algo:lstm}

\STATE MLP gender classification;\\
3.1 Concatenate $\textbf{v}_{i}$ and $\textbf{h}_{i}$ to be final features $\textbf{f}_{i}$: \\ $\textbf{f}_{i}=(\textbf{v}_{i}^{T},\textbf{h}_{i}^{T})^T$;\\
3.2 Input $\textbf{f}_{i}$ to MLP, get $G_{i}$:\\
$G_{i}=f(W\textbf{f}_{i}+b)$;
\label{algo:transfer}

\RETURN $G_i$;
\end{algorithmic}
\end{algorithm}

\noindent In Algorithm \ref{algo}, step 1 corresponds to micro-blog posts preprocess in Fig.~\ref{framework}. In this step we get virtual document matrix $\textbf{x}_i$ and virtual document vector $\textbf{v}_i$ of each user, and we also get document matrix $\textbf{x}_k$ and document vector $\textbf{v}_k$ of each source domain data. Step 2 corresponds to our sentiment representation learning. More specifically, step 2.1 illustrates the method of making a new source domain data, that is calculating the average similarity of each $\textbf{v}_k$ and all ${\textbf{v}_i}$, and select high similarity source domain data to be new source domain data. Step 2.3 is getting the lstm middle layer output to be our sentiment representation. Step 3 is the gender classification which uses a MLP model.

\noindent In Algorithm \ref{algo}, we calculate  the similarity, train a LSTM model, extract lstm layer output, and train a MLP to predict gender. Training a LSTM model has the computational complexity of $O(I\times W\times(H\times H+H\times V))$\cite{complexity}, where $I$ is the number of training epochs, $W$ is the number of tokens in the training set, $H$ is the size of hidden layer, and $V$ is the size of the vocabulary. Training a MLP model has the complexity of $ O(n\times m \times h\times h \times c \times i)$, where $n$ is the number of samples, $m$ is the feature dimension, $h$ is the number of hidden layer neuron, $i$ is the number of training epochs, and $c$ is the number of output class. The final complexity is their sum.

\section{Experiments}

\subsection{Dataset and Evaluation Measure}
All micro-blog data\footnote{https://github.com/WUT-IDEA/SRL-MLP/data/micro-blog} used in this paper are from the campaign of SMP technical evaluation launched by Chinese information society of China and Social media processing Specialized Committee, and the data set was provided and arranged by Sina micro-blog and Research Center of computing and information retrieval, Harbin Institute of Technology. But only these data are not enough, we need to learn the sentiment representation of micro-blog, and there is no label about sentiment in these data. We selected JD commodity review data\footnote{https://kexue.fm/archives/3863} with pos/neg labels as the source domain data because reviews have more pronounced sentiment polarities. Then we learn about the sentiment representation and transfer to the target domain of micro-blog. The way we get our JD commodity reviews are from the Internet. The data are collected and arranged by Jianlin Su, and they are open sourced. All the data we used in this paper can be found at our github\footnote{https://github.com/WUT-IDEA/SRL-MLP}.

\noindent Our evaluation measure is accuracy, which is the ratio of rightly classified number to total test samples number. For evaluating the overall performance of the model, we do 5-fold cross validation, and get their average accuracy.

\noindent Flowchart in Fig.~\ref{preprocess} shows the data preprocess. Social media text data have many characteristics that impede us analysing documents, such as short, non-standard, mixed of different advertisement, redundancy, etc. Micro-blog limits each text 140 words and for gender classification it is too short. We need larger corpus, so we link all posts that belongs to same user together to form a virtual document of micro-blog. Then we have to clean the data, including segmentation, deleting stop words, hyper links, special characters, and vectorize the words. We use word2vec\cite{w2v_1}\cite{w2v_2} to get each word's vector, and add every word vector up and get their average to be virtual document vector.

\subsection{Comparison of Different Document Representation}
In this part we try to find the best document representation. After getting virtual documents, we have only 3138 users labeled data, and the rate of male to female is 3 to 1. We have tried multiple ways to represent every user's virtual document, such as TF-IDF of all the words, TF-IDF of 256 gender differentiated words, 20 topics LDA, average word2vec which every word is represented by a 100 dimension vector. These 256 words are got by counting the words in male and female documents respectively, selecting 500 words that appear most, and eliminating the words that appear both. Our MLP model has only 2 hidden layer with 50 and 10 neurons respectively, and a dropout layer which drops neurons randomly by rate 0.4 and follows the first hidden layer. We put these document representation into MLP separately and make an comparison. Results in Tab.~\ref{doc_representation} tell us that word2vec is the best document representation for our problem. After getting document representation, we put them into classifiers to classify gender in the next subsection.

\begin{table}[t]
\centering
\caption{Accuracy of different document representations with MLP.}
\label{doc_representation}
\begin{tabular}{cc}
\hline
Word representation  & Accuracy(\%) \\ \hline\hline
TF-IDF               & 82.71  \\
Keywords TF-IDF      & 80.49  \\
LDA                  & 79.15  \\
Average word2vec     & \textbf{84.20}  \\
Average word2vec + LDA & 84.01  \\\hline
\end{tabular}
\end{table}

\subsection{Comparison of Different Gender Classifiers}
\begin{table}[t]
\centering
\caption{Accuracy of different classifiers.}
\label{comparison_of_classifiers}
\begin{tabular}{cc}
\hline
Classifiers             & Accuracy(\%) \\ \hline\hline
Logistic Regression     & 67.06 \\
Random Forest           & 72.15 \\
Support Vector Machine  & 76.34 \\
MLP                     & \textbf{84.20} \\
CNN                     & 74.21 \\
LSTM                    & 73.67 \\ \hline
\end{tabular}
\end{table}

In this part we try several traditional classifiers and some deep neural network classifiers to get preliminary experiment results on gender classification. Researchers have found a variety of classifiers, and each of them has its application field. We have tried several classifiers shown in Tab.~\ref{comparison_of_classifiers} to find the best suitable model, and results refer that MLP is the the best, better than CNN and LSTM, Logistic Regression model performs the poorest. Random Forest gets it highest accuracy when max split size is 400, random feature selecting size is set 100, and the number of trees is 15. Its highest accuracy is 72.15\%, a little higher than Logistic Regression. The Logistic Regression and Random Forest model are from Mahout. Compared with deep learning models, Support Vector Machine can get high-performance when data amount is small, here we use LibSVM\cite{libsvm} to classify gender. LibSVM gets the best accuracy when the svm type is set nu-SVC and the kernel is set linear.

\noindent Our LSTM model architecture is same with the second part of Fig.~\ref{framework}. Our CNN model has 32 filters, its kernel size width is 5 and length is 100 which is same with word dimension, then followed by an average pooling layer, finally flatted to link to a 50 neuron dense layer. As deep learning models, CNN and LSTM models have outstanding performance in text classification, but experimental results show that they are not very suitable for this gender classification problem. We apply them to sentiment analysis, and get effective results.

\subsection{Sentiment polarity features}
Before we extract our sentiment representation, we conduct an experiment to figure out how sentiment polarity features effect the result of gender classification. Our sentiment polarity features are 2 dimensional features. One is the virtual document sentiment polarity, the other is the rate of positive polarity. More specifically, we train a LSTM model on high similarity source domain data, then we input the virtual document matrix to LSTM model, get the output of virtual document sentiment polarity to be our first sentiment polarity feature. And we consider utilizing the sentiment polarity of each post, and getting the positive polarity rate $r_i$ of each virtual document to be our second sentiment polarity feature. $r_i=\frac{1}{a}\sum_{j=1}^{a}I(Polarity(p_{i,j})=1)$, in this formula, $a$ is the number of posts posted by a certain user, and $I$ is the indication function. When the polarity of post $p_{i,j}$ is positive, the value is 1; when polarity of post $p_{i,j}$ is negative, the value is 0. The result of adding these two sentiment polarity features are shown in Tab.~\ref{polarity_feature}. We can see from the figure that the sentiment polarity features have little improvement on the gender classification accuracy.

\begin{table}[t]
\centering
\caption{Cross validation accuracy(\%) of adding sentiment polarity features.}
\label{polarity_feature}
\setlength{\tabcolsep}{1.5mm}{
\begin{tabular}{cccccccccc}
\hline
epochs & 100 & 150 & 200 & 250 & 300 \\ \hline\hline
D1     & 82.13 & 82.85 & 80.86 & 81.65 & 81.33  \\
D2     & 86.12 & 84.44 & 85.48 & 85.72 & 84.13  \\
D3     & 85.96 & 83.24 & 84.21 & 82.69 & 83.89  \\
D4     & 83.09 & 83.33 & 83.09 & 81.81 & 83.01  \\
D5     & 84.52 & 84.44 & 84.76 & 84.04 & 85.00  \\ \hline
avg    & \textbf{84.36} & 83.66 & 83.68 & 83.18 & 83.47  \\ \hline
\end{tabular}}
\end{table}

\subsection{Data Imbalance}
\begin{table}[t]
\centering
\caption{Cross validation accuracy(\%) of before smote and after}
\label{fcnn_smote}
\setlength{\tabcolsep}{1.5mm}{
\begin{tabular}{cccccccccc}
\hline
\multicolumn{2}{c}{epochs} & 60 & 80 & 100 & 150 & 200 & 250 & 300 \\ \hline\hline
         & D1    & 81.10 & 83.41 & 83.25 & 80.62 & 82.93 & 82.05 & 82.53  \\
         & D2    & 83.41 & 84.29 & 84.68 & 83.97 & 83.65 & 82.69 & 82.13  \\
Before   & D3    & 83.97 & 86.20 & 84.76 & 83.57 & 84.05 & 84.29 & 83.89  \\
         & D4    & 82.93 & 82.85 & 82.53 & 83.73 & 82.61 & 82.21 & 82.21  \\
         & D5    & 84.04 & 84.28 & 83.33 & 83.41 & 84.12 & 81.42 & 80.95  \\ \hline
         & avg   & 83.09 & \textbf{84.20} & 83.71 & 83.06 & 83.47 & 82.53 & 82.34  \\ \hline
         & D1    & 82.76 & 81.20 & 85.01 & 83.56 & 86.08 & 83.56 & 84.63  \\
         & D2    & 84.42 & 84.04 & 83.56 & 86.61 & 85.76 & 85.59 & 84.68  \\
After    & D3    & 84.36 & 85.76 & 85.59 & 85.22 & 85.70 & 86.93 & 87.47  \\
         & D4    & 83.88 & 85.27 & 87.09 & 87.20 & 85.22 & 82.11 & 86.56  \\
         & D5    & 83.70 & 85.09 & 85.41 & 85.84 & 86.69 & 85.36 & 85.14  \\ \hline
         & avg   & 83.82 & 84.27 & 85.33 &85.69 & \textbf{85.89} & 84.71 & 85.70  \\ \hline
\end{tabular}}
\end{table}

We have only 3138 users labeled data, and the sample size rate of male to female is 3 to 1. For solving imbalance, we exploit smote oversampling\cite{smote} to generate data. The main idea of smote is to analyze the minority samples and add new minority samples to the data set. More specifically, smote proposes to find the nearest k minority sample $x_i,i=1,2,...,k,$ of minor sample $x_{old}$, then get the new minority sample $x_{new}=x_{old}+\sigma\frac{1}{k}\sum_{i=1}^{k}(x_i-x_{old})$. In this formula $\sigma$ is a random number from 0 to 1. Because MLP and LSTM need different formation of data input, we do oversampling separately. After smote, we can see that the accuracy improved by 1.69\%, and the results are shown in Tab.~\ref{fcnn_smote}. Now we have improved the accuracy from 84.20\% to 85.89\%.

\subsection{Performance of Our SRL-MLP}

\begin{figure}[h]
\centerline{\includegraphics[width=5cm,height=3.5cm]{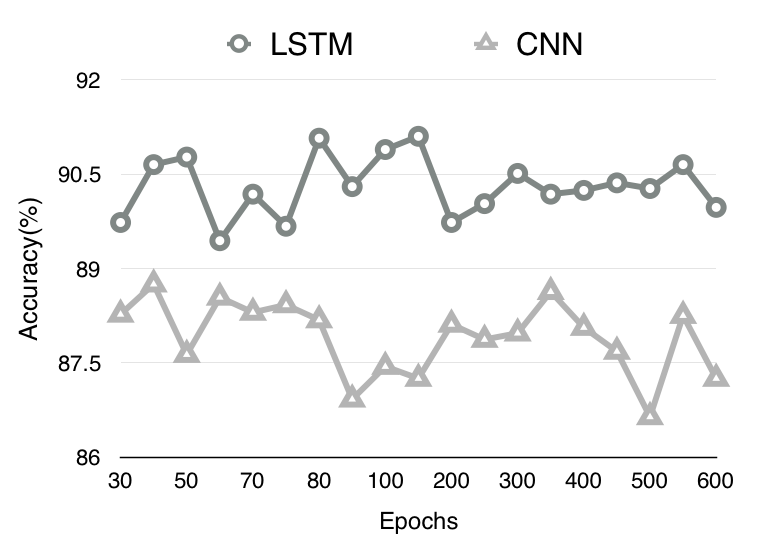}}
\caption{LSTM and CNN sentiment classification performance.}
\label{lstm_and_cnn}
\end{figure}

\noindent\textbf{sentiment analyzer selection.} For choosing a sentiment analyzer, we consider LSTM and CNN, and we make a comparison between them on entire JD reviews. Results in Fig.~\ref{lstm_and_cnn} tell us LSTM is more suitable for sentiment analysis than CNN in this problem.

\begin{figure}[t]
\begin{minipage}{1\linewidth}
  \centerline{\includegraphics[width=11.0cm, height=4.0cm]{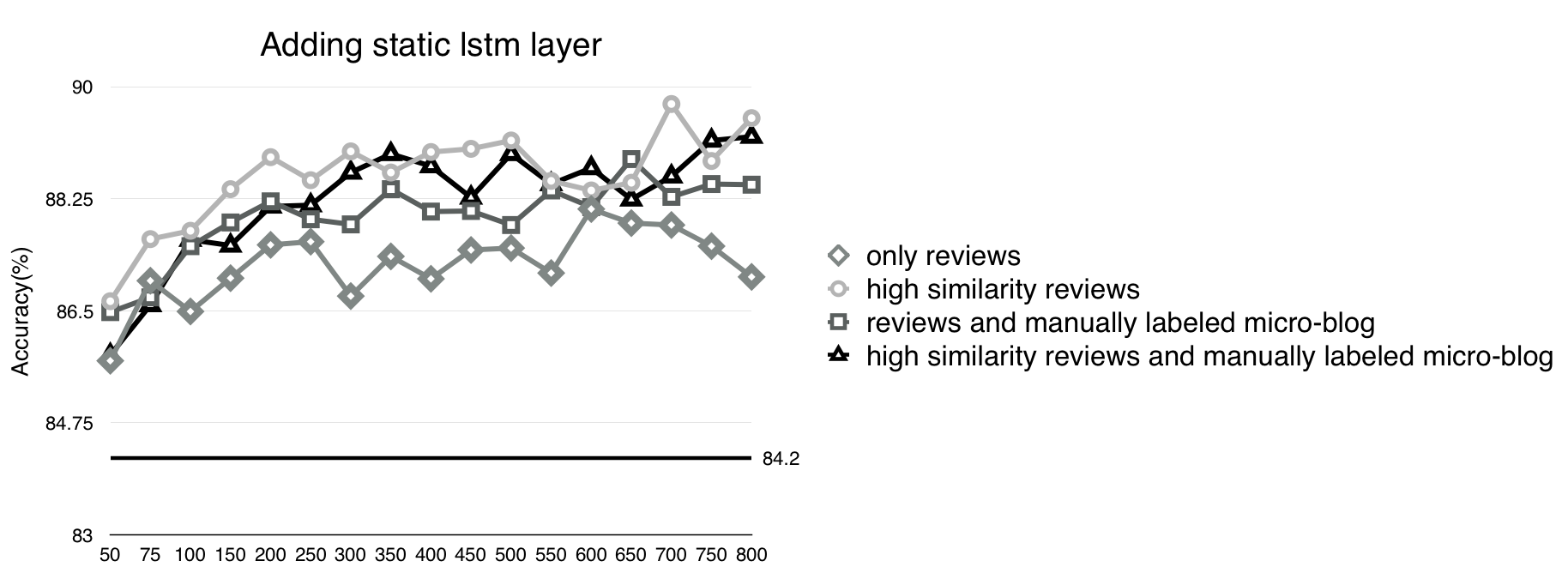}}
  \centerline{(1) Add frozen lstm layer.}
\end{minipage}
\vfill
\begin{minipage}{0.5\linewidth}
  \centerline{\includegraphics[width=6.0cm, height=4.0cm]{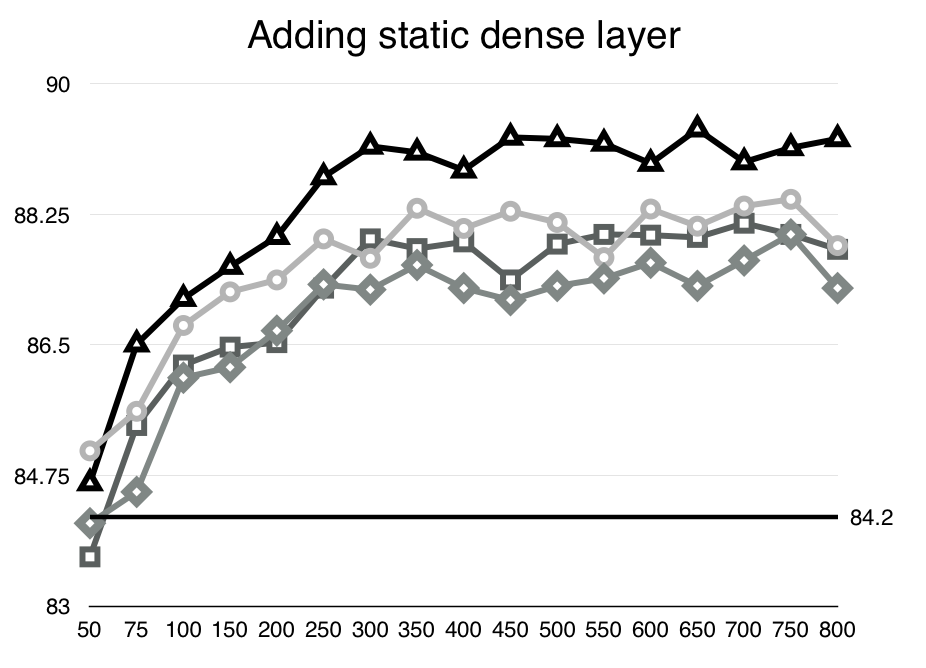}}
  \centerline{(2) Add frozen dense layer.}
\end{minipage}
\hfill
\begin{minipage}{0.5\linewidth}
  \centerline{\includegraphics[width=6.0cm, height=4.0cm]{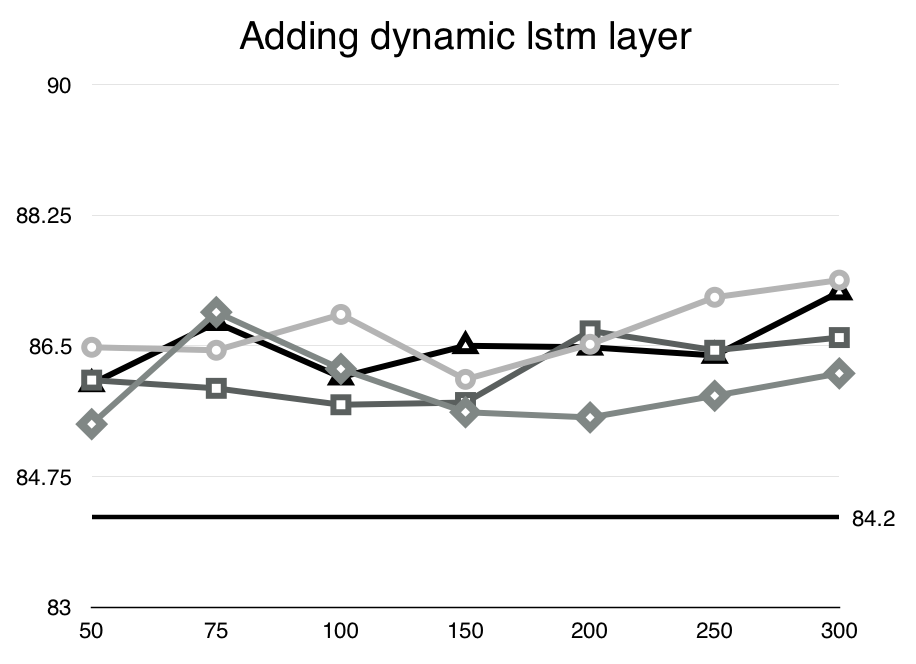}}
  \centerline{(3) Add finetuned lstm layer.}
\end{minipage}

\caption{This figure shows different sentiment representation learning results, that are selecting different source domain data and extracting different middle layer. In this figure, we draw a black horizontal line which represents the baseline.}
\label{result_group}
\end{figure}

\begin{table}[h]
\centering
\caption{Maximum accuracy(\%) of different source domain data training and different sentiment feature extraction.}
\label{sentiment_features}
\begin{tabular}{ccc}
\hline
train LSTM on               & added features          & accuracy(\%) \\
\hline\hline
                            & frozen lstm           & 88.09    \\
entire JD reviews           & frozen dense          & 87.98    \\
                            & finetuned lstm        & 86.95    \\
                  \hline
                            & frozen lstm           & \textbf{89.73}    \\
high similarity JD reviews  & frozen dense          & 88.45    \\
                            & finetuned lstm        & 87.38    \\

                  \hline
entire JD reviews           & frozen lstm           & 88.87    \\
and                         & frozen dense          & 88.13    \\
manually labeled micro-blog & finetuned lstm        & 86.75    \\
                  \hline
high similarity JD reviews  & frozen lstm           & 89.31    \\
and                         & frozen dense          & 89.26    \\
manually labeled micro-blog & finetuned lstm        & 87.22    \\
                  \hline
\end{tabular}
\end{table}

\noindent\textbf{Sentiment representation learning results.} After getting a suitable sentiment analyzer, we start learning sentiment representation. There is two steps for our sentiment representation learning: a)select source domain data; b)extract middle layer.

\noindent a) select source domain data. In this step we choose different corpus to train our LSTM. We make a accuracy comparison when LSTM sentiment analyzer was trained on 1)only JD reviews; 2)high similarity JD reviews; 3)JD reviews and manually labeled micro-blog posts; 4)high similarity JD reviews and manually labeled micro-blog posts. When source domain data are selected, we use Cosine similarity calculation method, and select the data whose average similarity can exceed 0.25, according to our preliminary parameter study, to be our new source domain data. When we extract the frozen lstm layer sentiment representation, the gender classification results are shown in Fig.~\ref{result_group}(1). We can see from the figure that, when LSTM is trained on high similarity data, the classification result is the best.

\noindent b) extract middle layer output. In this step we extract middle layer output to be our sentiment representation. We make a sentiment representation comparison between extracting 1)frozen lstm layer; 2)frozen dense layer; 3)finetuned lstm layer. We extract these middle layer outputs to get sentiment representation. The gender classification results are shown in Fig.~\ref{result_group}(1)-(3). In Fig.~\ref{result_group} we draw a black horizontal line to represent the baseline. We can see from this figure that, almost all sentiment representations have improved the classification accuracy. For a more direct viewing, their maximum is shown in Tab.~\ref{sentiment_features}.

\noindent From Tab.~\ref{sentiment_features}, we can see that after adding sentiment representation, accuracy of gender classification is much improved. Ideally, finetuning will increase more than frozen parameters, but due to small sample size, finetuning does not play a big role, and it only increases 1.49\%. But when training LSTM on high similarity data, and transfering frozen lstm layer sentiment representation, improvement is made from 85.89\% to 89.73\%, that is 3.84\% improvement.

\section{Conclusions and Future Work}
In this paper, we combine user profile with sentiment analysis. More specifically, we learn the representations of sentiment, combine the sentiment representations with virtual document vectors to form concatenated features, and put concatenated features into the user gender classification model to train a gender classifier. After introducing sentiment features into MLP, we improve classification accuracy by 5.53\%, from 84.20\% to 89.73\%. The transferred features are usually rich in expressiveness and effectiveness. This transferability can assist traditional feature extraction work, avoiding the time-consuming and complex nature of manual extraction of features.

In the future, we will try to introduce sentiment features into the classification of other user profile labels, such as age, education, occupation, etc. We will also try other possible effective features and try to mine the user's dynamic characteristics, such as interest, personality, etc.

\bibliographystyle{splncs04}
\bibliography{reference}

\end{document}